\def\year{2022}\relax

\documentclass[letterpaper]{article} 
\usepackage{aaai21}  
\usepackage{times}  
\usepackage{helvet} 
\usepackage{courier}  
\usepackage[hyphens]{url}  
\usepackage{graphicx} 
\usepackage{natbib}  
\usepackage{caption} 
\usepackage[ruled,vlined, linesnumbered]{algorithm2e}
\usepackage{booktabs}
\usepackage{tabularx}
\usepackage{array}
\usepackage{colortbl}
\newcolumntype{C}[1]{>{\centering\let\newline\\\arraybackslash}p{#1}}
\newcolumntype{M}[1]{>{\centering\let\newline\\\arraybackslash}m{#1}}

\usepackage{graphicx}
\usepackage{amsmath,amssymb} 
\usepackage{color}
\usepackage{multirow}
\usepackage[dvipsnames]{xcolor}
\usepackage[switch]{lineno}

\definecolor{lightskyblue}{rgb}{0.94,1.0,1.0}
\definecolor{lightgreen}{rgb}{0.94,1.0,0.94}
\definecolor{whitesmoke}{rgb}{0.92,0.92,0.92}
\definecolor{seashell}{rgb}{1.0,0.96,0.93}

\newcommand\ie{\textit{i.e.}}
\newcommand\eg{\textit{e.g.}}
\newcommand\viz{\textit{viz.}}

\title{Exploiting Negative Learning for Implicit Pseudo Label Rectification in Source-Free Domain Adaptive Semantic Segmentation}
\author{
    Xin Luo, Wei Chen, Yusong Tan, Chen Li, Yulin He, Xiaogang Jia \\
}

\affiliations{
    College of Computers, National University of Defense Technology
}

\begin{document}

\maketitle

\begin{abstract}

It is desirable to transfer the knowledge stored in a well-trained source model 
onto non-annotated target domain in the absence of source data. 
However, state-of-the-art methods for source free domain adaptation (SFDA) 
are subject to strict limits: 
1) access to internal specifications of source models is a must; and 
2) pseudo labels should be clean during self-training, 
making critical tasks relying on semantic segmentation unreliable. 
Aiming at these pitfalls, this study develops a domain adaptive solution 
to semantic segmentation with pseudo label rectification (namely \textit{PR-SFDA}), 
which operates in two phases: 
1) \textit{Confidence-regularized unsupervised learning}: 
Maximum squares loss applies to regularize the target model to ensure the confidence in prediction; and 
2) \textit{Noise-aware pseudo label learning}: 
Negative learning enables tolerance to noisy pseudo labels in training, 
meanwhile positive learning achieves fast convergence. 
Extensive experiments have been performed on domain adaptive semantic segmentation benchmark, \textit{GTA5 $\to$ Cityscapes}. 
Overall, \textit{PR-SFDA} achieves a performance of 49.0 mIoU, 
which is very close to that of the state-of-the-art counterparts. 
Note that the latter demand accesses to the source model's internal specifications, 
whereas the \textit{PR-SFDA} solution needs none as a sharp contrast.


\end{abstract}

\section{Introduction}












Unsupervised Domain Adaptation (UDA) aligns different domains and 
transfers the knowledge learned in well-annotated source domain 
onto non-annotated target domain. 
UDA plays an vital role in medical healthcare, autonomous driving and other scenarios 
where annotated data are far from sufficient for supervised learning. 
In UDA, the alignment between source and target data helps 
pre-trained source models to produce better predictions in target domain. 
The employment of UDA avoids supervised learning on target data, 
which can otherwise be obscured by the need of large amounts of data with fine-grained annotations.

\begin{figure}[tbp]
  \centering
      \includegraphics[width=\linewidth]{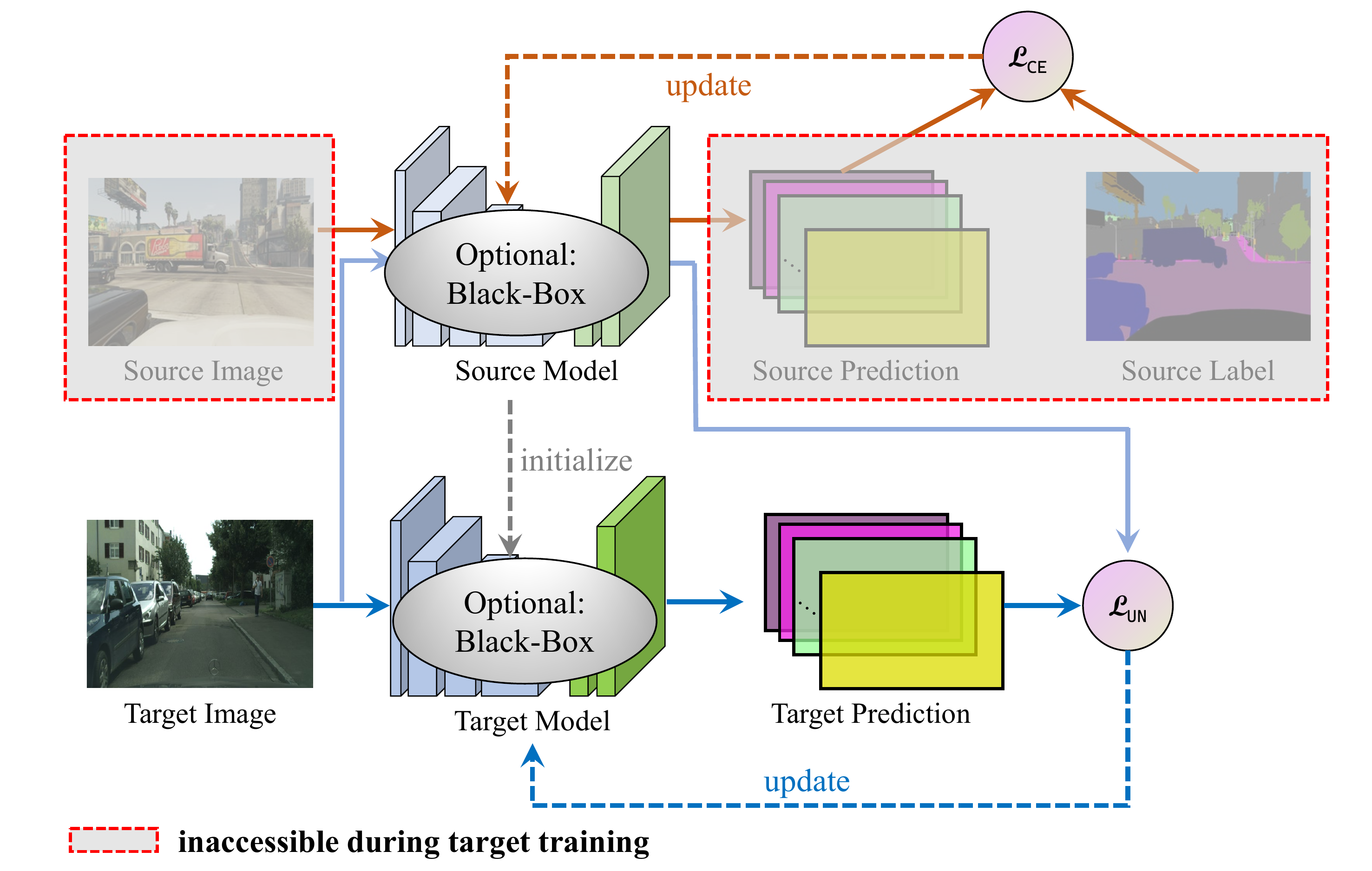}
  \caption{\textbf{Overview of Source-Free Domain Adaptation.} Typical SFDA methods have the access to \textit{a priori} knowledge for source models. In this study, we further limit the availability of model specifications. Our approach is model-independent, which enables even \textit{black-box} model's adaptation.}
  \label{fig:sfda}
\end{figure}

Typical UDA methods map input data from both domains into the same feature space. 
Early methods achieve this via minimizing the statistical discrepancy between source and target features. 
Recently, with the success of Generative Adversarial Networks \cite{nips2014gan}, 
many methods utilize adversarial learning to match and align the distribution of features. 
However, since these methods inevitably require the coexistence of source and target data, 
they cannot be applied into some source-absent scenarios, \ie, medical healthcare and financial prediction. 
Under these circumstances, source data are not available due to privacy concerns. 
Source-Free Domain Adaptation (SFDA) is suitable for these scenarios: 
(1) SFDA methods are not dependent on source data; 
(2) source models are always available in these scenarios, 
from which SFDA methods can extract domain knowledge. 
The SFDA theory has proved powerful 
for source data absent transfer learning in many research hotspots, 
\eg, image classification, object detection and semantic segmentation etc., 
which forms the basis of this study.

Recent advances in SFDA have enabled the simulation of source distribution. 
These methods extract statistics of specific layers in source models 
and try to match target distribution with them. 
Some other methods divide source models into feature extractors and classifiers. 
They fix the parameters of classifiers and optimize extractors to adapt for target data. 
However, all these methods require the detailed specification of source models. 
Practically, such \textit{a priori} knowledge about source models is not always accessible. 
Some methods benefit from self-training to increase accuracy in target domain,
but few of them explicitly deal with the inevitable noise in pseudo labels. 
In addition, since source models make main contributions to the quality of pseudo labels, 
source domains are supposed to provide models with strong capability of generalization. 
Nonetheless, most of existing SFDA methods focus on the adaptation, 
ignoring the importance of robust source models. 
In summary, an appropriate adaptation method is needed 
to take advantages of source models even without their specifications, 
and it should be robust enough to learn from noisy pseudo labels. 
To this end, there exists a pressing need for an approach to the pending problems:

\begin{itemize}
    \item without any \textit{a priori} knowledge about source models, 
    it is non-trivial to adapt towards target domain by mimicking source distribution, 
    or it will be prohibitively hard to selectively optimize part of the model. 
    This problem has long been underestimated.
    \item the employment of self-training is already a common practice 
    to enhance models on target non-annotated data. 
    Due to the intrinsic noises in pseudo labels, 
    enhancing the performance on target domain needs 
    to bridge the gap between the latest self-training methods 
    and an appropriate solution to rectifying noisy pseudo labels.
\end{itemize}

The strategy of this study is to 
first train a robust model in source domain 
then to adapt this model to target domain 
without the access to source data and source \textit{a priori}. 
With style augmentation upon source data, 
source models are enhanced towards slight data variance. 
The detection and rectification of pseudo labels will benefit the adaptation on target data. 
This study develops a source-free domain adaptation solution 
with pseudo label rectification (\textit{PR-SFDA}) operating in two phases:

\begin{itemize}

    \item \textit{confidence-regularized unsupervised learning} 
    for adaptation without \textit{a priori} knowledge of source models: 
    the application of probability regularization benefits target outputs 
    with higher prediction confidence. 
    The adaptation is guided towards regions with higher predictive entropy, 
    and the whole process does not need any explicit \textit{a priori} knowledge about source domain. 

    \item \textit{noise-aware pseudo label learning} 
    for enhanced self-training on target domain: 
    collaborative learning on positive labels 
    and negative labels extends naive self-training with noise rectification. 
    The utilization of negative learning enables 
    the tolerance to noisy pseudo labels, 
    whereas classical positive learning promotes fast convergence and accurate supervision.

\end{itemize}

Experiments on domain adaptive semantic segmentation have been carried out 
to evaluate the performance of \textit{PR-SFDA}. 
Experiments have also been performed to examine its effectiveness 
in pseudo label rectification 
and to investigate its sensibility to hyper-parameters.

The main contributions of this study are as follows:

\begin{itemize}
    \item this study develops a training method 
    for generating robust source models with class balance. 
    This enhances performance on long-tailed classes and 
    bridges the style gap between domains without access to target data.

    \item a complete solution has been fostered 
    to adapt a well-trained source model towards non-annotated target data. 
    It is noted that the adaptation is conducted 
    in the absence of both source data and layer specifications of source models.

    \item a collaborative learning method has been developed 
    to combine the advantages of positive and negative learning. 
    This brings both fast model convergence and effective noise rectification of pseudo labels.
\end{itemize}

\section{Related Work}

Numerous attempts have been made 
to improve the performance of domain adaptive semantic segmentation, 
which transfer the segmentation knowledge learned from annotated source domain 
to non-annotated target domain. 
Studies undertaken for this purpose focus on 
(1) minimizing the distribution discrepancy between domains and/or 
(2) improving the performance with unsupervised or semi-supervised methods. 
The most salient works along this direction are introduced as the follows:

\paragraph{Adversarial Learning for Domain Adaptive Segmentation.} 
FCNs \cite{arxiv16fcn} and ASN \cite{cvpr2018adaptseg} proposed frameworks 
to support cross-domain segmentation based on adversarial learning. 
By means of the game between feature extractor and domain discriminator, 
they achieved domain-invariant feature extraction. 
Upon these methods, ADVENT \cite{cvpr2019advent} proposed 
the concept of adversarial entropy minimization, 
which utilized entropy as certainty metric 
to regularize the model for domain-consistent confident predictions. 
Similarly, MSL \cite{iccv2019msl} studied the imbalanced gradients of entropy loss between easy and hard samples. 
They proposed Maximum Squares Loss to prevent the training process being dominated by easy samples. 

\paragraph{Self-training for Domain Adaptive Segmentation.}
CBST \cite{eccv2018cbst} and CRST \cite{iccv2019crst} proposed 
novel UDA frameworks based on self-training, \ie, 
optimization via alternatively updating pseudo labels and target model. 
Their studies employ class-balanced loss and confident constraints 
to further regularize self-training. 
Following them, FDA \cite{cvpr2020fda} employed 
the mean predictions of multiple models to regularize self-learning.

CCM \cite{eccv2020ccm} evaluated the quality of pseudo labels 
via spatial contextual layout similarity, 
dropping noisy samples that have inconsistent contextual layouts. 
IntraDA \cite{cvpr2020intrada} separated target domain into easy and
hard splits and applied intra-domain adaptation to transfer hard samples. 
DAST \cite{aaai2021dast} introduced discriminator attention to concentrate on potential noisy regions. 
MRNet \cite{ijcai2020mrnet, ijcv2021seguncertainty} detected and rectified noisy pseudo labels 
by means of scale variance in target predictions. 
Recently, ProDA \cite{cvpr2021proda} exploited the feature distances 
from prototypes to enable online correction of pseudo labels, 
achieving tremendous performance advantage over previous methods.

\paragraph{Source-Free Domain Adaptive Segmentation.}
To adapt source knowledge in the absence of source data, 
UBNA \cite{arxiv2020ubna} studied the normalization layer statistics for adaptation. 
This enabled them to mix the statistics from both domains. 
Similarly, SFDA \cite{arxiv2021sfda} employed a generator 
to mimic the distribution of source features, 
which came from the adversary between fake-source and target batch normalization statistics.

This study aims at domain adaptive semantic segmentation and 
has the following major concerns: 
(1) to adapt and transfer source knowledge 
without access to any explicit source knowledge 
(neither source data/feature nor source model specification), and meanwhile 
(2) to detect and rectify noisy pseudo labels 
only with the predictions from \textit{black-box} source models.


\section{Preliminaries}
This section clarifies: 
(1) the notations and operations of SFDA, which forms the basis of this study; 
(2) the basic concept of \textit{Negative Learning}, which contributes to pseudo label rectification.

\subsection{Source-Free Unsupervised Domain Adaptation}

Under typical UDA settings, 
a labeled source dataset $\mathcal{D}_s=\{(x_s^i,y_s^i)\}_{i=1}^{N_s}$, 
an unlabeled target dataset  $\mathcal{D}_t=\{x_t^i\}_{i=1}^{N_t}$ and 
a well-trained source model $\mathcal{M}_s$ work collaboratively 
to enhance model $\mathcal{M}$'s performance on target domain, where 
$x_s$, $x_t$ denote source and target images, 
$y_s$ is source annotation and $N_s$, 
$N_t$ denote the number of source and target samples, respectively.

Equation~\ref{eq:loss_da} formulates the basic optimization objective 
for source-present domain adaptation (SDA), where 
$\mathcal{L}_{SRC}$ is supervised loss in source domain, 
$\mathcal{L}_{DA}$ is the adaptation loss in both domain 
(which comes from adversarial learning or other training techniques) and 
$\mathcal{L}_{TGT}$ is regularization loss on target domain.

\begin{equation}
\label{eq:loss_da}
\mathcal{L}_{SDA}=\mathcal{L}_{SRC}+\mathcal{L}_{TGT}+\mathcal{L}_{DA}
\end{equation}

In the absence of source data, \ie, $\mathcal{D}_s$, 
only a well-trained source model and non-annotated target data are available. 
Under this setting, source-free domain adaptation (SFDA) methods are applied.

For SFDA, source and target data are utilized separately in different domains. 
In source domain, the model is optimized with $\mathcal{L}_{SRC}$ 
to provide a well-trained source model $\mathcal{M}_s$. 
In target domain, the model is optimized with $\mathcal{L}_{TGT}(\mathcal{M}, \mathcal{M}_s;x_t)$, 
without the existence of source data. 

\subsection{Negative Learning}

\begin{figure}[tbp]
  \centering
      \includegraphics[width=\linewidth]{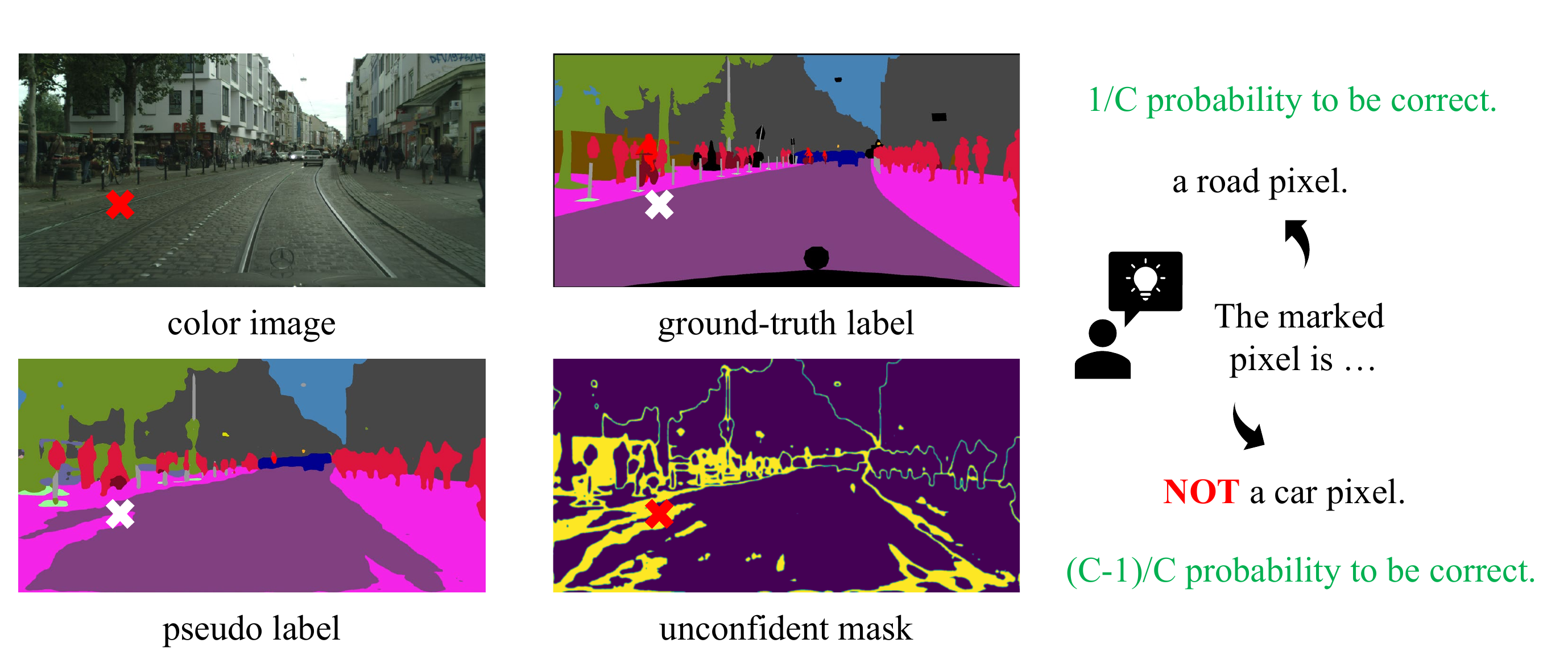}
  \caption{Negative Learning}
  \label{fig:concept_of_negative_learning}
\end{figure}

Given an input as well as its label, 
cross-entropy loss updates the model in a \textbf{positive} manner 
by maximizing the prediction probability towards the corresponding class. 
This is defined as positive learning. 
When labels are noisy, naive positive learning will lead to error accumulation. 
To alleviate this, negative learning updates models differently. 
Negative learning tries to minimizing the prediction probability 
that the given input \textbf{does not} belong to the corresponding label.

Figure~\ref{fig:concept_of_negative_learning} illustrates the concept of negative learning. 
The marked pixel has a less confident pseudo label, 
which is noisy (labeled as \textit{sidewalk} instead of \textit{road}). 
For simplicity, we assume that all the labels are subject to the uniform distribution. 
Thus, the probability that the pixel is labeled correctly is only $\frac{1}{C}$. 
By contrast, negative learning uses its complementary label, whose probability of correctness is $\frac{C-1}{C}$. 
In this way, negative learning can help avoid the impact of noisy labels 
via providing correct information even from wrong labels.

The loss calculation for positive and negative learning 
is defined in Equation~\ref{eq:pl} and Equation~\ref{eq:nl}, respectively.

\begin{equation}
    \mathcal{L}_{PL} =-\sum_{c=1}^{C}{y_c\log p_c(x)}
    \label{eq:pl}
\end{equation}

\begin{equation}
    \mathcal{L}_{NL} =-\sum_{c=1}^{C}{\overline y_c\log (1-p_c(x))}
    \label{eq:nl}
\end{equation}

Based upon these preliminaries, this study proposes \textit{PR-SFDA}, 
a domain adaptive solution for semantic segmentation, aiming at 
(1) adapting domain knowledge in the absence of source data and 
(2) rectifying intrinsic noises in pseudo labels 
without the help of source model \textit{a priori}.

\section{Method}


This section first demonstrates 
how to generate a class-balanced robust source model for SFDA scenarios. Then, it details 
the design and operation of the \textit{PR-SFDA} solution in two phases: 
(1) \textit{confidence-regularized unsupervised target adaptation}, and 
(2) \textit{noise-rectified target self-training}.

\subsection{Architecture of \textit{PR-SFDA}}

\begin{figure*}[tbp]
  \centering
      \includegraphics[width=0.8\linewidth]{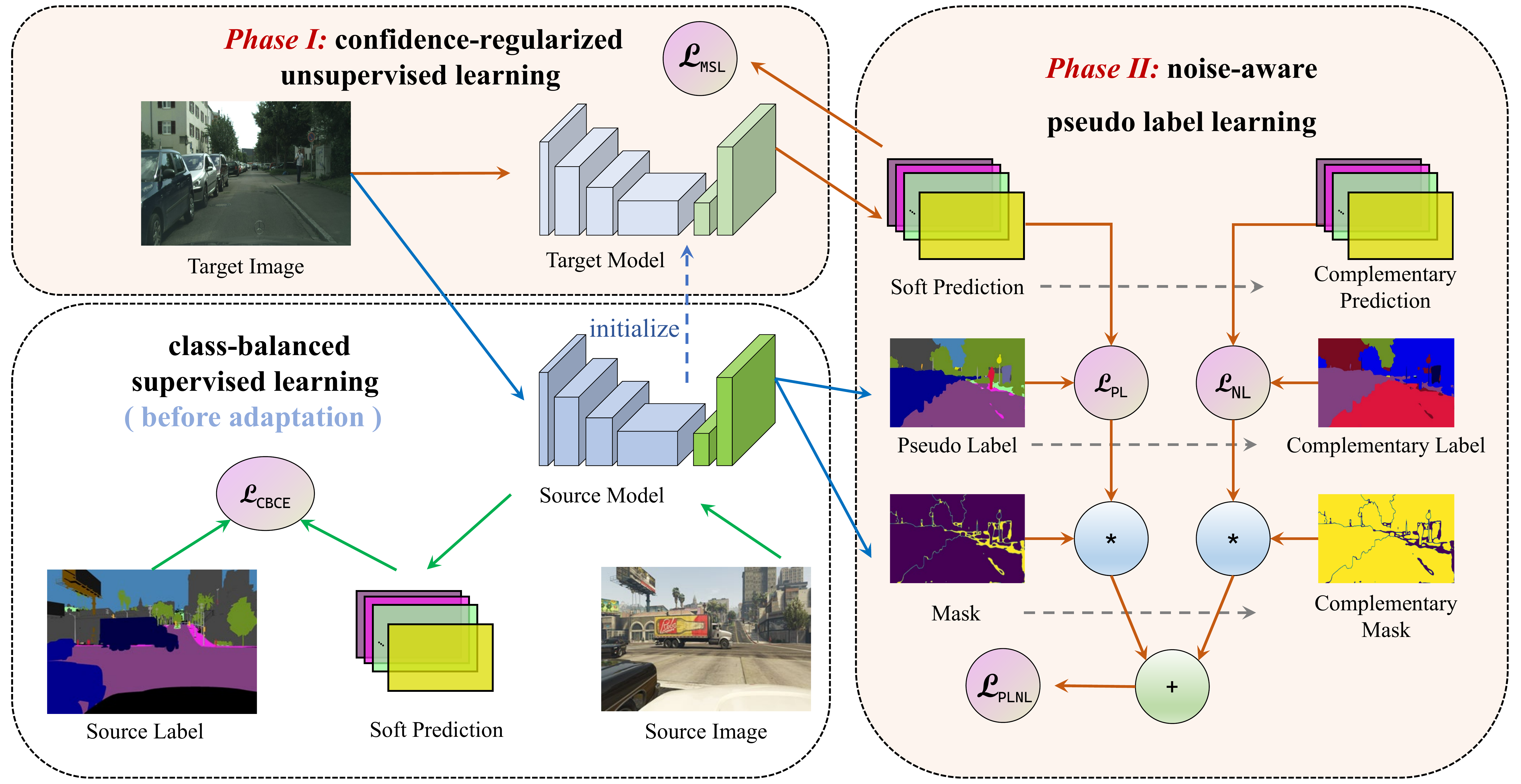}
  \caption{Overview of the Architecture and Operation of the \textit{PR-SFDA} Solution}
  \label{fig:arch_of_prsfda}
\end{figure*}

Figure~\ref{fig:arch_of_prsfda} gives an overview of the architecture and operation of \textit{PR-SFDA}. 
The \textit{PR-SFDA} solution operates in two phases: 
(1) confidence regularization helps improve the pre-trained source model for confident predictions. 
(2) the regularized model produces pseudo labels and confidence masks. 
Then, positive learning and negative learning work collaboratively 
to extend self-training with pseudo label rectification.

\subsection{Class-Balanced Supervised Learning}
\label{ch:phase0}

SFDA methods cannot access source data for direct supervision. 
This requires robust source models to produce reliable segmentation results for target data.

To alleviate class imbalance of source model, 
this study employs class-balanced cross-entropy loss for model optimization. 
Class-balanced cross-entropy (CBCE) loss uses class-aware weights 
to increase the contribution of those long-tailed rare classes, 
which helps improve the performance on them. 
CBCE loss is defined in Equation~\ref{eq:CBCE}, 
where $w_c$ is the weight for the $c$-th class.
Each weight is calculated according to 
the pixels' occurrence frequency among the whole dataset and their ratio in the image.

\begin{equation}
    \mathcal{L}_{\small{CBCE}}(p(x),y)=-\sum_{c=1}^{C}{w_c * y_c\log p_c(x)}
    \label{eq:CBCE}
\end{equation}

To bridge the style discrepancy between source images and unavailable target images, 
this study characterizes style variance between domains as noise perturbation. 
To this end, we apply color perturbation to augment source images. 
This enables abundant style variance and can slightly reduce the domain gap between source and target images.

The objective for source training can be formulated as Equation~\ref{eq:loss_src}.

\begin{equation}
\label{eq:loss_src}
    \mathcal{L}_{SRC}=\mathop{\mathbb{E}}\limits_{\small{(x_s,y_s)\sim \mathcal{X}_s \times \mathcal{Y}_s}}\mathcal{L}_{\small{CBCE}}(p(\widetilde{x_s})),y_s)
\end{equation}

\subsection{Confidence-Regularized Unsupervised Learning}
\label{ch:phase1}

In this phase, \textit{PR-SFDA} aims to adapt source model onto unlabeled target domain. 
Due to the significant discrepancy between source and target domains, 
source models are not guaranteed to generate accurate prediction for target images. 
Instead, the generated results often suffers from noisy regions, 
which have lower prediction confidence.

Entropy minimization has been widely adopted 
to penalize regions with higher entropy (\ie, lower confidence). 
Since easy samples often come with higher entropy, 
\textit{PR-SFDA} adopts maximum squares loss (MSL) \cite{iccv2019msl} as certainty regularization 
to prevent the training process being dominated by easy samples. 
$\mathcal{L}_{MSL}$ is defined as in Equation~\ref{eq:loss_msl}, 
which has balanced gradients for each class and enables confident predictions for target data.

\begin{equation}
    \label{eq:loss_msl}
    \mathcal{L}_{MSL}=-\mathop{\mathbb{E}}\limits_{\small{x_t\sim \mathcal{X}_t}} \frac{1}{2}\sum_{c=1}^C(p_c(x_t))^2
\end{equation}

After confidence regularization, \textit{PR-SFDA} is capable of 
predicting coarse-grained results in target domain. 
Previous studies employed self-training for further model optimization. 
These methods rely on either source representations or model specifications 
to rectify noises in pseudo labels. 
This limits their application in cloud-based environments and other scenarios 
where these \textit{a priori} knowledge might be unavailable. 
As a consequence, there exists a technical barrier 
with noise rectification in the absence of architecture specification.

\subsection{Noise-Aware Pseudo Label Learning}
\label{ch:phase2}

Current model in \textit{PR-SFDA} has been capable of generating coarse-grained results 
for target images (pseudo labels as well as confident maps). 
Based on confident maps, a pre-defined threshold (\eg, $0.6$ in our experiments) is utilized 
to generate binary invalid masks, 
which indicates the distribution of potential noisy regions . 

In this phase, \textit{PR-SFDA} extends self-training with rectification in those noisy regions, 
which is achieved via the collaboration of positive learning and negative learning. 
As illustrated in Figure~\ref{fig:concept_of_negative_learning}, 
negative learning assigns complementary labels for each prediction. 
This increases the probability that a supervision signal is correct. 
The generation of complementary labels is demonstrated as in Algorithm~\ref{alg:generate_clabel}.

\begin{algorithm}[h]
    \caption{Generation of Complementary Label.}
    \label{alg:generate_clabel}
    \KwIn{Pseudo Label $y \in [0,C]^{H\times W}$ }
    \KwOut{Complementary Label $\overline y \in [0,C]^{H\times W}$}
    \BlankLine
    $\overline{y}$=y.copy( )
    \BlankLine
    \ForEach{lab in $[0,C]$}{
        tmpLab = random.randint(0, C)
        \BlankLine
        \While{tmpLab == lab}{
           tmpLab = random.randint(0, C)
        }
        \BlankLine
        
        $\overline{y}$[y==lab]=tmpLab
    }
\end{algorithm}

For negative learning, \textit{PR-SFDA} calculates 
the complementary prediction, pseudo label and invalid mask from the original outputs. 
After this, positive and negative learning collaborate with each other, 
which results in $\mathcal{L}_{PLNL}$. $\mathcal{L}_{PLNL}$ is defined in Equation~\ref{eq:loss_nlpl}, where 
$M$ denotes invalid mask, 
$y$ denotes pseudo label, 
$\overline y$ denotes randomly shuffled pseudo label, 
$p$ denotes soft prediction for input image $x$ and 
$\otimes$ is element-wise multiplication.  

\begin{equation}
    \mathcal{L}_{PLNL}=(1 - M) \odot \mathcal{L}_{PL}+\lambda( M \odot \mathcal{L}_{NL})
    \label{eq:loss_nlpl}
\end{equation}

In summary, the \textit{PR-SFDA} solution aims 
to detect noisy predictions with less confidence and 
to alleviate invalid supervision from noisy regions via negative learning.

\section{Experiments}

\begin{figure*}[tbp]
    \centering
    \includegraphics[width=\textwidth]{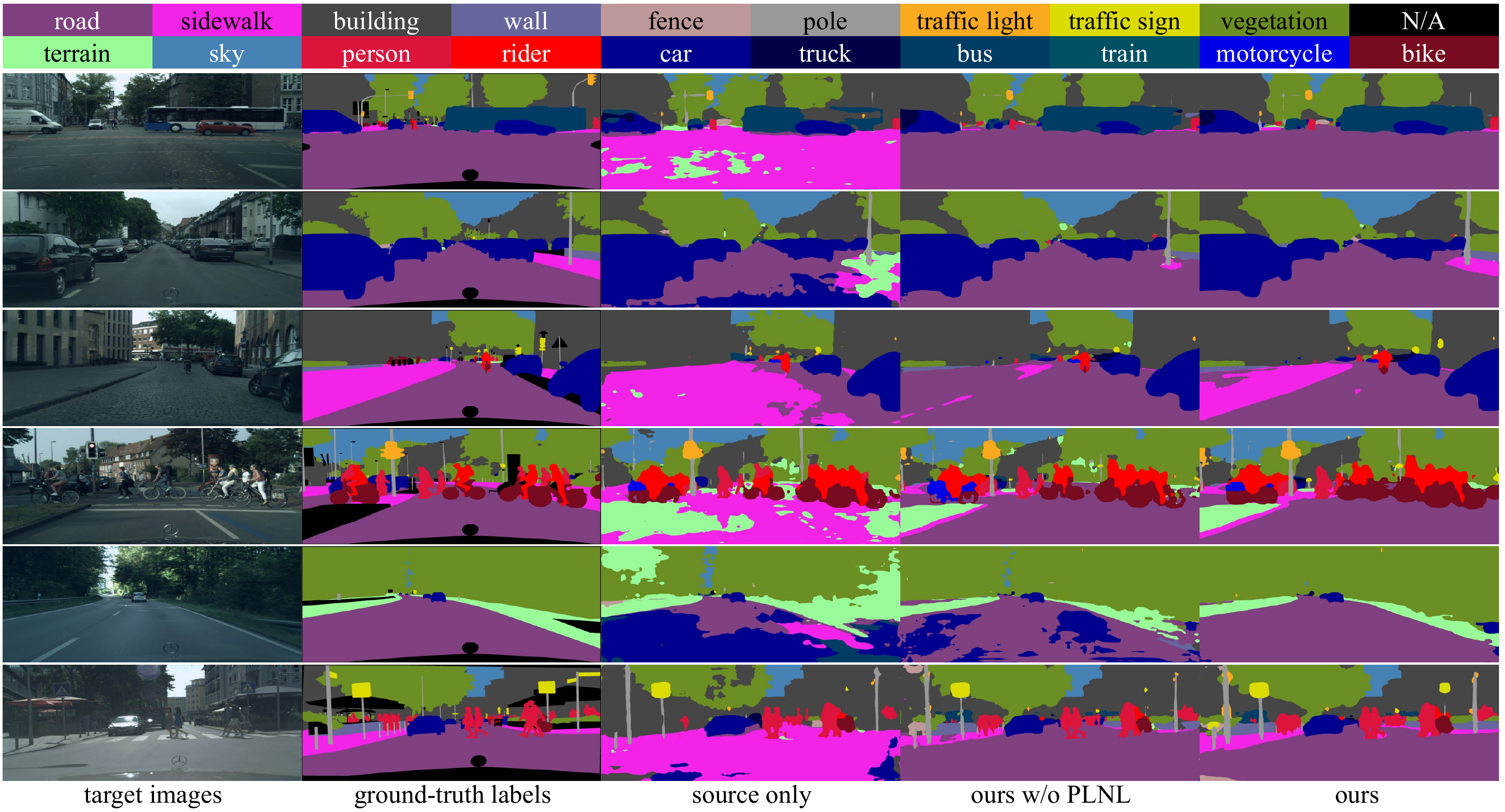}
    \caption{Qualitative Results on \textit{GTA5 $\to$ Cityscapes}}
    \label{fig:gta2cs_results}
\end{figure*}

\setlength\tabcolsep{2.8pt}
\begin{table*}[h!]
\footnotesize
\centering
\caption{\textbf{Performance on GTA5$\to$Cityscapes.} 
The tail classes are highlighted in \textcolor{blue}{blue}. 
The column \textbf{SF} denotes whether the adaptation method is source-free. 
For the listed methods, the results are taken from original papers. }
\begin{tabular}{ c | c | c c c c c c c c c c c c c c c c c c c | c }
\toprule[1pt]
Method  &  SF                             & \rotatebox{90}{road} & \rotatebox{90}{sidewalk\ } & \rotatebox{90}{building} & \rotatebox{90}{\textcolor{blue}{wall}} & \rotatebox{90}{\textcolor{blue}{fence}} & \rotatebox{90}{\textcolor{blue}{pole}} & \rotatebox{90}{\textcolor{blue}{light}} & \rotatebox{90}{\textcolor{blue}{sign}} & \rotatebox{90}{veg} & \rotatebox{90}{\textcolor{blue}{terrain}} & \rotatebox{90}{sky} & \rotatebox{90}{\textcolor{blue}{person}} & \rotatebox{90}{\textcolor{blue}{rider}} & \rotatebox{90}{car} & \rotatebox{90}{\textcolor{blue}{truck}} & \rotatebox{90}{\textcolor{blue}{bus}} & \rotatebox{90}{\textcolor{blue}{train}} & \rotatebox{90}{\textcolor{blue}{mbike}} & \rotatebox{90}{\textcolor{blue}{bike}} & \rotatebox{90}{\textbf{\textcolor{red}{mIoU}}} \\
\midrule[1pt]
Source Only & / & 38.0 & 13.7 & 79.5 & 14.8 & 22.3 & 32.3 & 33.8 & 30.3 & 80.8 & 14.8 & 75.4 & 60.0 & 28.9 & 56.4 & 39.1 & 32.5 & 0.2 & 29.6 & 37.0 & 37.9 \\
\midrule[0.5pt]
ASN &  & 86.5 & 25.9 & 79.8 & 22.1 & 20.0 & 23.6 & 33.1 & 21.8 & 81.8 & 25.9 & 75.9 & 57.3 & 26.2 & 76.3 & 29.8 & 32.1 & 7.2 & 29.5 & 32.5 & 41.4 \\
BDL &  & 91.0 & 44.7 & 84.2 & 34.6 & 27.6 & 30.2 & 36.0 & 36.0 & 85.0 & 43.6 & 83.0 & 58.6 & 31.6 & 83.3 & 35.3 & 49.7 & 3.3 & 28.8 & 35.6 & 48.5 \\
CLAN &  & 87.0 & 27.1 & 79.6 & 27.3 & 23.3 & 28.3 & 35.5 & 24.2 & 83.6 & 27.4 & 74.2 & 58.6 & 28.0 & 76.2 & 33.1 & 36.7 & 6.7 & 31.9 & 31.4 & 43.2 \\
ADVENT &  & 89.9 & 36.5 & 81.6 & 29.2 & 25.2 & 28.5 & 32.3 & 22.4 & 83.9 & 34.0 & 77.1 & 57.4 & 27.9 & 83.7 & 29.4 & 39.1 & 1.5 & 28.4 & 23.3 & 43.8 \\
IntraDA &  & 90.6 & 37.1 & 82.6 & 30.1 & 19.1 & 29.5 & 32.4 & 20.6 & 85.7 & 40.5 & 79.7 & 58.7 & 31.1 & 86.3 & 31.5 & 48.3 & 0.0 & 30.2 & 35.8 & 46.3 \\
APODA &  & 85.6 & 32.8 & 79.0 & 29.5 & 25.5 & 26.8 & 34.6 & 19.9 & 83.7 & 40.6 & 77.9 & 59.2 & 28.3 & 84.6 & 34.6 & 49.2 & 8.0 & 32.6 & 39.6 & 45.9 \\
FDA  &  & 92.1 & 51.5 & 82.3 & 26.3 & 26.8 & 32.6 & 36.9 & 39.6 & 81.7 & 40.7 & 78.2 & 57.8 & 29.1 & 82.8 & 36.1 & 49.0 & 13.9 & 24.5 & 43.9 & 48.8 \\
\rowcolor{lightgreen}
\textbf{PR-SFDA} & $\checkmark$ & 91.3 & 41.8 & 85.2 & 34.5 & 24.2 & 34.4 & 36.3 & 40.7 & 85.6 & 42.6 & 87.0 & 60.4 & 30.8 & 86.2 & 37.9 & 40.3 & 1.4 & 22.7 & 48.2 & 49.0 \\
\bottomrule[1pt]
\end{tabular}
\label{tab:sota_results}
\end{table*}

This section details experiments and results. 
Experimental studies were performed 
to examine the effectiveness of \textit{PR-SFDA}.
Specifically, we evaluated its performance
in domain adaptive semantic segmentation, 
using two typical benchmarks, \ie, 
\textit{GTA5 $\to$ Cityscapes} and \textit{SYNTHIA $\to$ Cityscapes}.


\subsection{Dataset and Metric} 
Specifically, GTA5 is a synthetic dataset rendered from video game engine. 
It contains 24,966 images with fine-grained semantic annotation. 
Cityscapes is a popular benchmark for urban-scene semantic segmentation, 
which collects street-view images from 50 cities. 
It contains a training split of 2,975 images, 
a validation split of 500 images and a testing split of 888 images.
SYNTHIA is another synthetic dataset. 
This study employed one of its subsets, \viz, SYNTHIA-RAND-CITYSCAPES.
It contains 9400 annotated synthetic images and its annotation is compatible with Cityscapes. 

This study adopted the established evaluation protocol from
previous work, calculating pre-class Intersection-over-Union (IoU) 
and mean IoU (mIoU) over all classes on the \textit{val} split of Cityscapes. 
Equation~\ref{eq:miou} is the definition of mIoU, 
where $p_{ij}$ denotes the number of pixels 
that belong to the \textit{i-th} class 
and are classified as the \textit{j-th} class.

\begin{equation}
    \label{eq:miou}
    \mathrm{mIoU}=\frac{1}{C}\sum_{i=1}^C{\frac{p_{ii}}{(\sum_{j=1}^C p_{ij}) + (\sum_{j=1}^C p_{ji}) - p_{ii} }}
\end{equation}

\subsection{Implementation Details} 
Our method is implemented using the DeepLab framework with the ResNet-101 backbone. 
The model is implemented on the Pytorch platform and runs on a single RTX 6000 GPU with 24GB memory. 
We train the whole model through back-propagation, 
The decoding layers are trained with a learning rate 10 times that of the pre-trained encoding layers. 
During source training, the whole network is optimized with the stochastic gradient descent (SGD) algorithm,
with a momentum of $0.9$ and a weight decay factor of $5 \times 10^{-4}$). 
During target adaptation, the optimization is performed via AdamW optimizer,
with $\beta$ set as  $(0.9, 0.99)$ and a weight decay factor of $5 \times 10^{-4}$). 
The initial learning rate is $2.0^{-4}$ and 
is scheduled using the polynomial decay with a power of $0.9$. 
For target regularization and self-training, the initial learning rate is initialized as $1.0^{-5}$.
During training, we resize the images to $1280 * 760$, $1024 * 512$ for synthetic and realistic datasets, respectively. 
The batch size is $2$ for all the experiments.

\subsection{Results of Adaptive Semantic Segmentation}

Figure~\ref{fig:gta2cs_results} shows the qualitative results 
on \textit{GTA5 $\to$ Cityscapes} adaptation benchmark. 
It is obvious that our method can significantly improve the baseline source-only model. 
The introduced target regularization (\ie, Maximum Squares Loss, MSL) helps 
improve the confusing segmentation (\eg, road and sidewalk). 
By comparing the right-most two columns in Figure~\ref{fig:gta2cs_results}, 
it can be noticed that the prediction can be significantly improved with the proposed \textit{PR-SFDA} solution, 
\ie, collaborative positive and negative learning.

Table~\ref{tab:sota_results} present quantitative results, 
indicating that the proposed method achieves close performances to many state-of-the-art methods, 
in spite of the absence of source data and source \textit{a priori}. 
The performance is even superior to some baseline source-dependent adaptation methods.

Furthermore, with weighted cross-entropy loss 
and collaborative positive and negative learning, 
\textit{PR-SFDA} shows superior performance on those long-tailed categories 
(highlighted as \textcolor{blue}{blue} in the table). 
The reason is two-fold. 
On the one hand, the weighted cross-entropy loss (in the \textit{``class-balanced supervised learning'' phase}) 
drives the source model to put emphasis on those hard classes during back-propagation, 
leading to increasing contribution of those classes. 
On the other hand, the prediction on those areas is generally less confident. 
In these regions, negative learning will be selectively applied (in the \textit{``noise-aware pseudo label learning'' phase}). 
This helps reduce the impact of false-prone pseudo labels in those regions 
and increase prediction confidence and accuracy.

\subsection{Ablation Study}

To validate the contributions of different phases in \textit{PR-SFDA}, 
we conduct some ablation experiments to compare performance under different settings. 
Table~\ref{tab:ablation_phases} shows the quantitative comparison amongst different phases.

\subsubsection{Impact of Source Augmentation}

In Table~\ref{tab:ablation_phases}, the difference between methods \textit{SO} and \textit{SO(AUG)} 
lies in the employment of data augmentation. 
For \textit{SO}, the naive images are used without augmentation, and 
\textit{AUG} denotes the employment of data augmentation 
(\ie, color perturbation in this study). 
The style variance between source and target images leads to domain shifts, 
which is the main cause of limited performance in target domain. 
Without the presence of both source and target images, 
it is non-trivial to bridge domain gaps via adversarial learning, style translation, etc. 
To this end, with color perturbation, \textit{PR-SFDA} augments source images 
to increase their style diversity. 
This helps extend the range of source domain, and thus can 
help reduce the domain discrepancy between two domains, 
leading to \textbf{+6.3\%} mIoU increase.

\setlength\tabcolsep{6.8pt}
\begin{table}[t]
  \caption{Ablation study of different phases}
  \centering
  \footnotesize
    \begin{tabular}{cccccc|c}
     \toprule[1pt]
     \small{SO}       & \small{AUG}      & \small{ENT}      &
     \small{MSL}      & \small{ST}       & \small{ST-NLPL}     & \textbf{mIoU} \\
     \midrule[1pt]
     $\checkmark$     &                  &                  &
                      &                  &                  & 37.9 \\
     $\checkmark$     & $\checkmark$     &                  &
                      &                  &                  & 40.3 \\
     $\checkmark$     & $\checkmark$     & $\checkmark$     &
                      &                  &                  & 43.2 \\
     $\checkmark$     & $\checkmark$     &                  &
     $\checkmark$     &                  &                  & 45.6 \\
     $\checkmark$     & $\checkmark$     &                  &
     $\checkmark$     & $\checkmark$     &                  & 46.2 \\
     $\checkmark$     & $\checkmark$     &                  &
     $\checkmark$     &                  & $\checkmark$     & 49.0 \\
     \bottomrule[1pt]
    \end{tabular}%
  \label{tab:ablation_phases}%
\end{table}%

\subsubsection{Impact of Target Regularization}

If source and target domains are subject to similar distributions, 
the pre-trained source model will perform well on target images. 
However, such assumption is not guaranteed during practice. 
This urges further optimization on target images. 
To regularize on target images without any annotations, 
we adopt the widely used uncertainty regularization. 
Specifically, we investigate two common uncertainty metric, 
\ie, minimum entropy (\textit{ENT}) and maximum squared prediction probability (\textit{MSL}). 
As shown in Table~\ref{tab:ablation_phases}, both \textit{SO(AUG)+ENT} and \textit{SO(AUG)+MSL} 
can improve the performance on target domain. 
However, the employment of MSL brings more promotion, 
which can be explained by its balanced gradients towards each class.   

\subsubsection{Impact of Pseudo Label Rectification}

Pseudo label learning can further regularize model in non-annotated target domain. 
As shown in Table~\ref{tab:ablation_phases}, 
with the help of self-training (\textit{ST}), 
\textit{SO(AUG)+MSL+ST} gains a considerable performance increase. 
However, naive self-training needs multiple rounds of iterative updating between pseudo labels and model, 
which is computationally expensive. 
Moreover, naive self-training sets up confidence thresholds to filter invalid pseudo labels. 
This will lead to a bias toward those easy classes, 
as long-tailed hard classes tend to be predicted with lower confidence. 
Table~\ref{tab:ablation_phases} compares \textit{PR-SFDA} (\textit{SO(AUG)+MSL+NLPL}) 
with naive self-training (\textit{SO(AUG)+MSL+ST}). 
The results demonstrate that \textit{PR-SFDA} reaches 
superior performance to naive self-training, 
demonstrating the effectiveness of pseudo label rectification in \textit{PR-SFDA}. 

\subsubsection{Trade-off between Positive and Negative Learning}

\setlength\tabcolsep{15pt}
\begin{table}[tbp]
  \centering
  \caption{Ablation study of $\lambda_{nl}$}
    \begin{tabular}{c|ccc}
    \toprule[1pt]
    $\lambda_{nl}$ & 0.1  &  0.5 & 1   \\
    \midrule[0.5pt]
    mIoU  & 48.5 & 48.9 & 49.0 \\
    \bottomrule[1pt]
    \end{tabular}%
  \label{tab:ablation_param}
\end{table}%

To make better trade-off between positive and negative learning, 
we also conduct ablation experiments 
to select the optimal coefficient for positive learning and negative learning. 
Table~\ref{tab:ablation_param} shows the quantitative results of \textit{PR-SFDA} 
with different coefficients for negative learning 
($\lambda_{nl}=[0.1, 0.5, 1.0]$). 
The table implies that a higher weight of negative learning 
brings slight performance increases. 
However, the table also shows that \textit{PR-SFDA} is 
generally not sensitive to the parameter $\lambda_{nl}$, 
with similar performances under different settings.

\section{Conclusions}

Aiming at the grand challenges for domain adaptive semantic segmentation 
in the absence of both source data and internal model specifications, 
this study developed a source free domain adaptation solution 
with pseudo label rectification (namely \textit{PR-SFDA}) for domain agnostic learning, 
consisting of two phases: 
(1) \textit{confidence-regularized unsupervised learning}, 
where different confidence regularizations have been investigated 
to enhance the predictive certainty over non-annotated target images. and 
(2) \textit{noise-aware pseudo label learning}, 
where a collaborative learning method has been developed 
to detect and rectify the intrinsic noises in pseudo labels.

Experimental results indicated that: 
(1) confidence regularization could guide the optimization 
towards those less confident regions, 
leading to a performance improvement of +13.2\% mIoU 
than the model trained with augmented source data; 
(2) without access to any explicit \textit{a priori} knowledge on source domain, 
\textit{PR-SFDA} could rectify pseudo labels for effective self-training, 
achieving a performance of 49.0 mIoU. 
It held potentials in domain agnostic learning and self-training 
with limited model/domain \textit{a priori} knowledge.


\end{document}